\documentclass{article} 
\usepackage{GEM_workshop_2024,times}

\usepackage{amsmath,amsfonts,bm}

\def\eqref#1{equation~\ref{#1}}

\def\1{\bm{1}}

\DeclareMathAlphabet{\mathsfit}{\encodingdefault}{\sfdefault}{m}{sl}
\SetMathAlphabet{\mathsfit}{bold}{\encodingdefault}{\sfdefault}{bx}{n}

\usepackage{hyperref}
\usepackage{url}
\usepackage{enumitem}

\usepackage{amsmath}
\usepackage{algorithm}
\usepackage{algorithmic}
\usepackage{subcaption}
\usepackage{wrapfig}
\usepackage[pdftex]{graphicx}
\usepackage[rightcaption]{sidecap}
\usepackage[textsize=scriptsize]{todonotes}

\title{Generative Humanization \\ for Therapeutic Antibodies}

\author{Cade Gordon\thanks{Equal contribution}  \hspace{.1mm} \thanks{Work performed while at BigHat Biosciences} \\
University of California, Berkeley\\
\And
Aniruddh Raghu$^{*}$\\
BigHat Biosciences\\
\And
Peyton Greenside\\
BigHat Biosciences\\ 
\And 
Hunter Elliott\\ 
BigHat Biosciences\\
\texttt{helliott@bighatbio.com}
}

\newcommand{\cutsubsectionup}{\vspace{-0.2cm}}

\iclrfinalcopy 
\begin{document}

\maketitle

\begin{abstract}
Antibody therapies have been employed to address some of today's most challenging diseases, but must meet many criteria during drug development before reaching a patient. \textit{Humanization} is a sequence optimization strategy that addresses one critical risk called immunogenicity --- a patient's immune response to the drug --- by making an antibody more `human-like' in the absence of a predictive lab-based test for immunogenicity. However, existing humanization strategies generally yield very few humanized candidates, which may have degraded biophysical properties or decreased drug efficacy. Here, we re-frame humanization as a conditional generative modeling task, where humanizing mutations are sampled from a language model trained on human antibody data. We describe a sampling process that incorporates models of therapeutic attributes, such as antigen binding affinity, to obtain candidate sequences that have both reduced immunogenicity risk \textit{and} maintained or improved therapeutic properties, allowing this algorithm to be readily embedded into an iterative antibody optimization campaign. We demonstrate \textit{in silico} and in lab validation that in real therapeutic programs our generative humanization method produces diverse sets of antibodies that are both (1) highly-human and (2) have favorable therapeutic properties, such as improved binding to target antigens.
\end{abstract}

\section{Introduction}
\cutsubsectionup
Antibodies are the fastest growing drug class, with approved molecules treating a breadth of disorders ranging from cancer to autoimmune disease to infectious disease \citep{carter2018next}. Many candidate therapeutic antibodies are derived from non-human e.g., murine or camelid sources, and modern antibody formats such as multi-specifics or antibody-drug conjugates can require heavy sequence engineering after discovery. This increases the risk of immunogenicity, where Anti-Drug Antibodies (ADAs) result in either fast clearance of the drug or adverse events \citep{hwang2005immunogenicity}. While antibody sequence humanness is only roughly correlated with immunogenicity, humanization is widely employed to decrease immunogenicity risk \citep{prihoda2022biophi}. However, even targeted humanizing changes may alter drug efficacy, and any such engineering must be undertaken while maintaining or improving the ``drug-likeness'' (developability) and intended function of the resulting antibody.

Canonically, humanization is a time-consuming, manual or semi-manual ``cut-and-paste" process requiring specialized expertise. 
Machine learning-based approaches to humanization have been proposed to address this challenge -- 
many of these \citep{marks2021humab, prihoda2022biophi, ramon2023abnativ}
use models built on large antibody datasets \citep{kovaltsuk2018observed} to suggest specific mutations which greedily optimize humanness. Published methods employ various mitigating measures aimed at preserving antibody function such as avoiding changes in the Complementarity Determining Regions (CDRs), which are key for target binding, or relying on structural features \citep{choi2015antibody, tennenhouse2023computational}(related work in Appendix \ref{sec:app_related_work}). However these methods may still alter antibody function \citep{chirino2004minimizing, harding2010immunogenicity}, and generally yield one or a small number of humanized candidates which, if unsuccessful, can prevent the drug from advancing in development.

In this work, we propose re-framing humanization as a conditional generative sequence modeling task, and outline a new humanization algorithm that, given a starting antibody, generates \textit{multiple} humanized candidates enriched for therapeutic properties of interest.  When this method is combined with a modern high-throughput wet lab, it allows reliable and efficient humanization that can be performed either in a single step, or incrementally during an iterative optimization campaign: at each round of optimization, we generate many humanized sequences enriched for properties of interest, validate them quickly in the lab, and select top performers  for further development in the next round.

From a technical standpoint, our humanization algorithm operates by introducing controlled, humanizing mutations to antibody sequences. These mutations are obtained by \textit{sampling} from a masked language model (MLM) trained on human antibody sequences. Sampling from an MLM allows us to generate many more diverse, human-like sequences than what existing (mostly deterministic) methods produce. 
We outline a product of experts formulation for sampling that incorporates 
oracle models trained to predict therapeutic attributes of antibodies, such as binding affinity. This allows us to generate \textit{multiple} humanized candidates that maintain (or even improve) the therapeutic properties of the starting antibody.

In experiments, we first evaluate our algorithms \textit{in silico}, demonstrating we can obtain a large number of highly human candidates with favorable therapeutic attributes. Secondly, we conduct lab validation and demonstrate that in two real therapeutic programs, our method generates humanized antibodies that have improved binding to a target antigen as compared to baselines.
\section{Methods}
\subsection{Background}
\textbf{Antibody humanization.} 
To be a viable therapeutic candidate, an antibody must be ``developable'' - \textit{e.g.}, it must be synthesizable, thermostable, and of specific relevance here, non-immunogenic \citep{jarasch2015developability}. 
Humanization aims to address this requirement, by taking an engineered antibody and altering its amino acid sequence such that it resembles human antibodies more closely, while preserving its structure, function, and developability. 

\textbf{Notation.} Denote the set of 20 amino acids by $\mathcal{A}=\{r_0, r_1,\ldots,r_{19}\}$. 
Let $\mathbf{x} = [x_0, \dots, x_{L-1}]\in \mathcal{A}^L$ denote an antibody, representing a sequence of $L$ amino acids with each $x_i \in \mathcal{A}$.

Let $f_k: \mathcal{A}^L \mapsto \mathbb{R}$ be a scoring function, or \textit{oracle}, that evaluates an antibody $x$ on some metric of interest. Examples include binding affinity $f_{K_{D}}$, melting temperature $f_{T_m}$, and humanness $f_h$.

Further, let $m : \mathcal{A}^L \mapsto \mathbb{R}^{L \times 20}$ denote a \textit{masked language model} (MLM) that takes an input antibody $\mathbf{x}$ (potentially with some residues replaced by \texttt{<MASK>}) and outputs a ${L \times 20}$ matrix $\mathbf{Z} = [\mathbf{z}_0, \ldots, \mathbf{z}_{L-1}]^\top$, where each \mbox{$\mathbf{z}_l$} contains log probabilities over the set of residues in $\mathcal{A}$. The probability distribution over the residues at location $l$ is obtained using the \texttt{softmax} function with temperature $\tau$, giving the probability of each residue $r_i$ at location $l$ as:
\begin{align}
  p(r_i) = \texttt{softmax}(\mathbf{z}_l/\tau)[i] = \frac{\exp(\mathbf{z}_l[i]/\tau)}{\sum_{k=1}^{20} \exp(\mathbf{z}_l[k]/\tau)}. \label{eqn:softmax}
\end{align}
\subsection{Humanization via Sampling}
\label{sec:methods_sampling}
\textbf{Concept.} We humanize a starter antibody by mutating it over a series of steps. Each step's mutation is obtained by \textit{sampling} from the output distribution of a masked language model (MLM) conditioned on the current sequence. Importantly, the MLM is trained on a large human antibody dataset (The Observed Antibody Space, OAS \citep{kovaltsuk2018observed}). Samples from such an MLM will tend to introduce more human-like mutations, which can be confirmed with an orthogonal humanness metrics such as the OASis percentile \citep{prihoda2022biophi}.

Our algorithm proceeds as follows. 
We first propose a starting antibody sequence $\mathbf{x}^{(0)}$, mutable residue locations $\mathbf{I}$, and a sampling temperature $\tau$. 
For each step $j$ of the iterative humanization, we pass the sequence $\mathbf{x}^{(j)}$ through the MLM $m$ (potentially after masking), generating a matrix of log probabilities $\mathbf{Z}$. We take $\mathbf{z}_{\mathbf{I}[j]}$, the log probabilities at location $\mathbf{I}[j]$, and sample from the resulting distribution following \eqref{eqn:softmax}. The mutated location $\mathbf{I}[j]$ is infilled with the sampled residue, resulting in $\mathbf{x}^{(j+1)}$. This process then repeats until all indices $\mathbf{I}$ are filled.

\textbf{On the choice of masking strategy.} The simplest variant is to not mask at all -- \textit{Unmasked Sampling}. 
If we mask all of $\mathbf{I}$ at the start and progressively infill, we obtain \textit{Autoregressive Denoising Sampling} (similar to Autoregressive Diffusion Models \citep{hoogeboom2021autoregressive}).  If at each iteration $j$, we only mask $x_{\mathbf{I}[j]}$, we obtain \textit{Gibbs-like Sampling}. We compare these different strategies in our experiments. Appendix \ref{sec:app_methods} has full details for all three variants. 


\textbf{On \texttt{argmax} Humanization.} Our algorithm samples from the MLM output distribution at mutable locations, rather than infilling mutable locations with the residue that has the maximal MLM output (i.e., taking an \texttt{argmax}). This choice results in many more humanized candidates (Section \ref{sec:expts:1}). \texttt{argmax} infilling (possibly over multiple rounds) without masking gives the Sapiens algorithm \citep{prihoda2022biophi}, a baseline with which we compare with in experiments. We also define two new variants on Sapiens that do incorporate masking:
\begin{itemize}[nosep]
    \item Random Masking Argmax: Mask the mutable input residues, pass the sequence through the MLM, infill all masked locations with the \texttt{argmax} over the MLM outputs.
    \item Iterative Masking Argmax: For each mutable residue, mask it out, pass the sequence through the MLM and infill using the \texttt{argmax} operator. Repeat for all mutable residues.
\end{itemize}

\subsection{Guided Sampling for Attribute-Aware Iterative Humanization}
One important drawback with the approach described so far is that the process is completely agnostic to important functional or developability attributes of the starting sequence such as strong binding affinity and thermostability. The humanized candidates can therefore have substantially worse properties along these axes, as we demonstrate in Section \ref{sec:expts:1}. Furthermore, it assumes that humanization will be undertaken as a single step at either the beginning or end of a campaign, rather than as part of an iterative optimization process.

To address this, we modify the sampling to enrich for attributes of therapeutic importance while iteratively improving humanness.
Assume access to a set of oracles $\{f_k\}_{k=1}^{K}$, each of which scores an antibody $\mathbf{x}$ based on some attribute. 
Then, at each mutable location $l$, instead of sampling from the MLM distribution, we sample from a product of experts (PoE) \citep{hinton2002training} distribution.
We define $\mathbf{s}_{k,x_l} \in \mathbb{R}^{20}$ to be a vector of scores that oracle $f_k$ produces for all 20 possible mutations of $x_l$. Interpreting these scores as log probabilities, the PoE distribution at this location has the following log probability of residue $r_i$:
\begin{align}
\log p_{\text{PoE}}(r_i) = \mathbf{z}_l[i]/\tau + \sum_{k=1}^K \mathbf{s}_{k,x_l}[i]/\tau_k  -\log Z, \label{eq:poe}
\end{align}
where $\mathbf{z}_l$ are the MLM's log probabilities at location $l$, $\tau_k$ represents each oracle's temperature, and $Z$ is a normalizing constant. Here, high probability is assigned to a residue $r_i$ that has both: (1) high likelihood under the MLM, representing `humanness';  and (2) high scores under the oracles, representing other desired therapeutic attributes, weighted by the temperature terms.
By sampling from \mbox{$r' \sim \texttt{softmax}(\mathbf{z}_{\mathbf{I}[j]}/\tau + \sum_{k=1}^K \mathbf{s}_{k,x_{\mathbf{I}[j]}}/\tau_k)$} instead of \eqref{eqn:softmax}, we obtain our updated algorithm: \textit{Guided Sampling Humanization}. In our experiments, we demonstrate that sampling in this way generates sequences which progressively improve both humanness \textit{and} desired therapeutic attributes.

\textbf{Comparison to PoE in existing work.} Each step of our guided sampling algorithm constructs a \textit{local PoE} distribution over mutations at a single location, which allows for tractable sampling. This implicitly assumes a notion of local functional smoothness in sequence space, which we show is effective in practice.
Prior work \citep{emami2023plug, hie2022high} has formulated PoE distributions over multiple-mutation trajectories; however, sampling from these distributions necessitates approximations since calculating the partition function exactly is intractable -- see Appendix \ref{sec:app_related_work}.

\section{Experiments}
\subsection{Experimental Setup}
\textbf{Datasets.} For Masked Language Model (MLM) training, following several earlier works \citep{ruffolo2021antiberty, prihoda2022biophi},  we use a dataset of 4 million sequences from Open Antibody Space \citep{kovaltsuk2018observed} (OAS). OAS data processing details are in the appendix.  
Oracle models were first pre-trained on in-house Next Generation Sequencing (NGS) datasets with between 500k and 1M sequences, and then fine-tuned to ~8k binding and thermostability measurements. 

\textbf{Model and training details.} We train a BERT-style \citep{devlin2018bert} masked language model (MLM) with $\sim$25 million parameters on the OAS subset described above. 
We train two oracles -- one to predict binding affinity, and one to predict melting temperature. Each of these is an ensemble of 1D CNNs with a ByteNet/CARP architecture \citep{yang2022convolutions} -- details in Appendix \ref{sec:app_expt_details}. 

\textbf{Humanness metrics.} We use an established humanness metric, the OASis percentile \citep{prihoda2022biophi}, which is well-correlated with immunogenicity risk (with higher humanness indicating lower immunogenicity). When assessing the contribution of experts to guided sampling, we also use the log likelihood of a sequence under the MLM as a continuous humanness proxy, which is correlated with the OASis percentile score (Appendix \ref{sec:app_expt_details}).

\begin{figure}[t]
    \centering
    \begin{subfigure}[b]{0.25\textwidth}
        \centering
        \includegraphics[width=\textwidth]{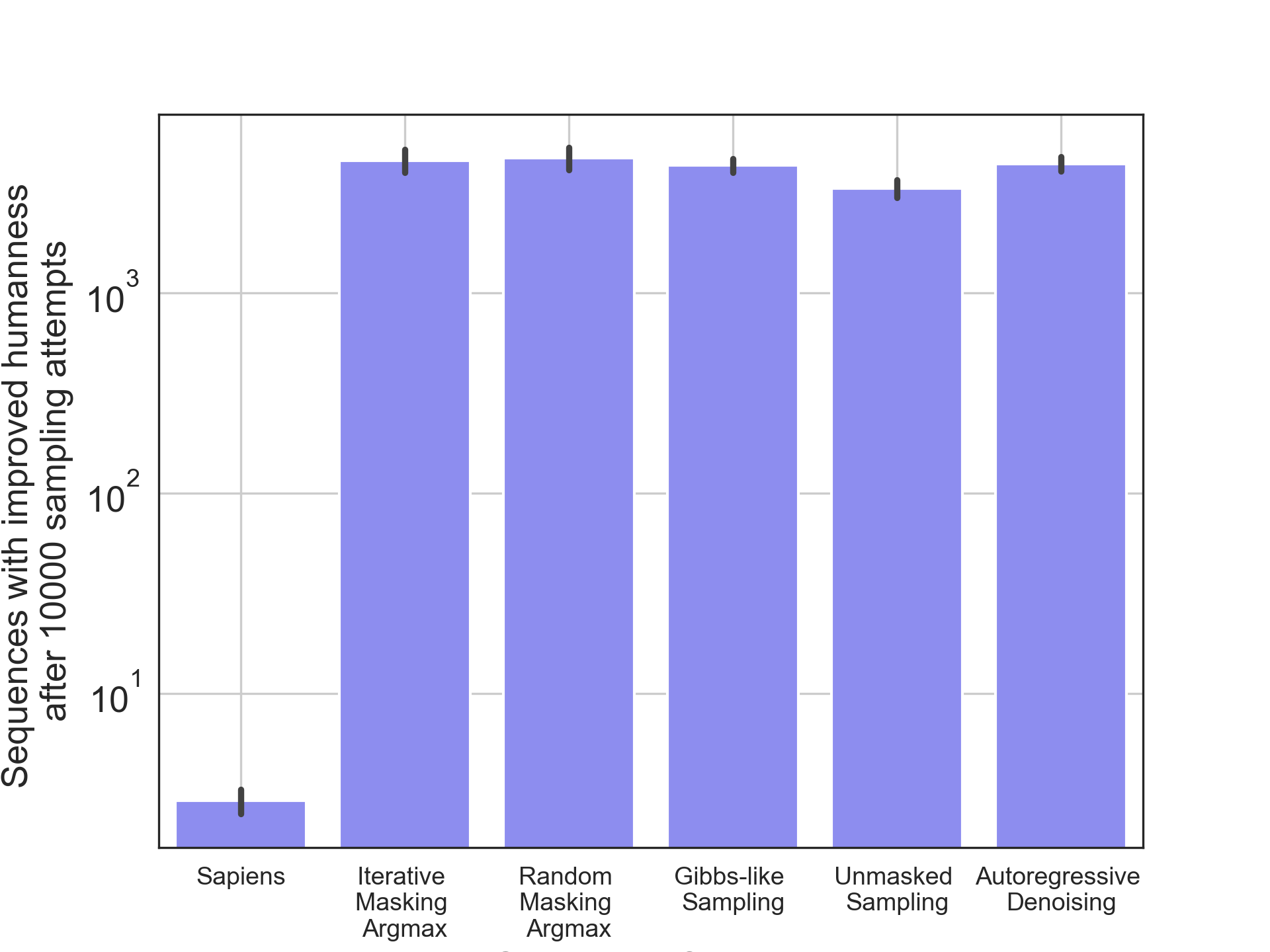}
        \caption{\scriptsize Multiple Murine Starters}
        \label{fig:fig2_murine}
    \end{subfigure}
    \begin{subfigure}[b]{0.25\textwidth}
        \centering
        \includegraphics[width=\textwidth]{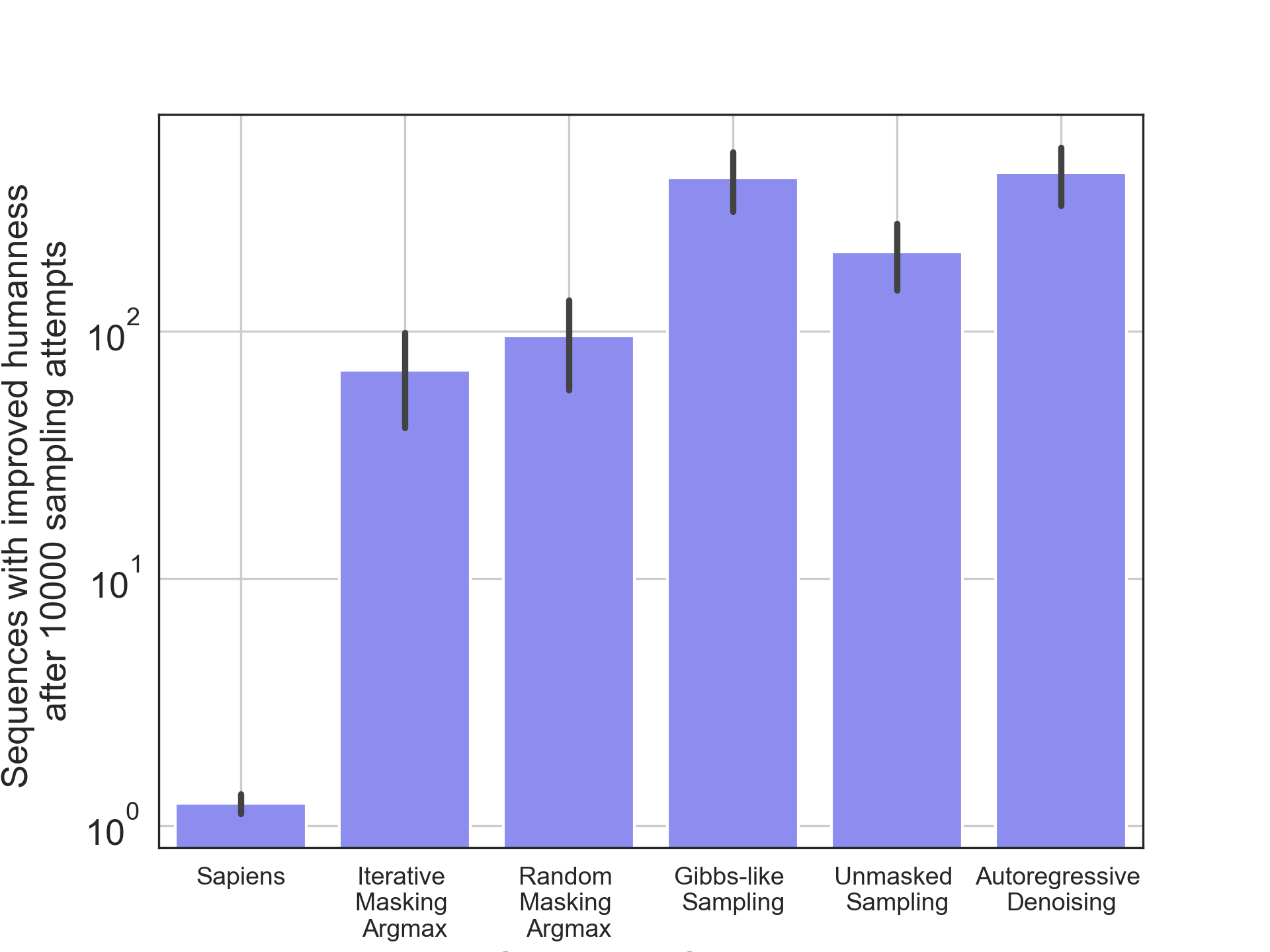}
        \caption{\scriptsize Multiple Human Starters}
        \label{fig:fig2_human}
    \end{subfigure}
    \begin{subfigure}[b]{0.22\textwidth}
        \centering
        \includegraphics[width=\textwidth]{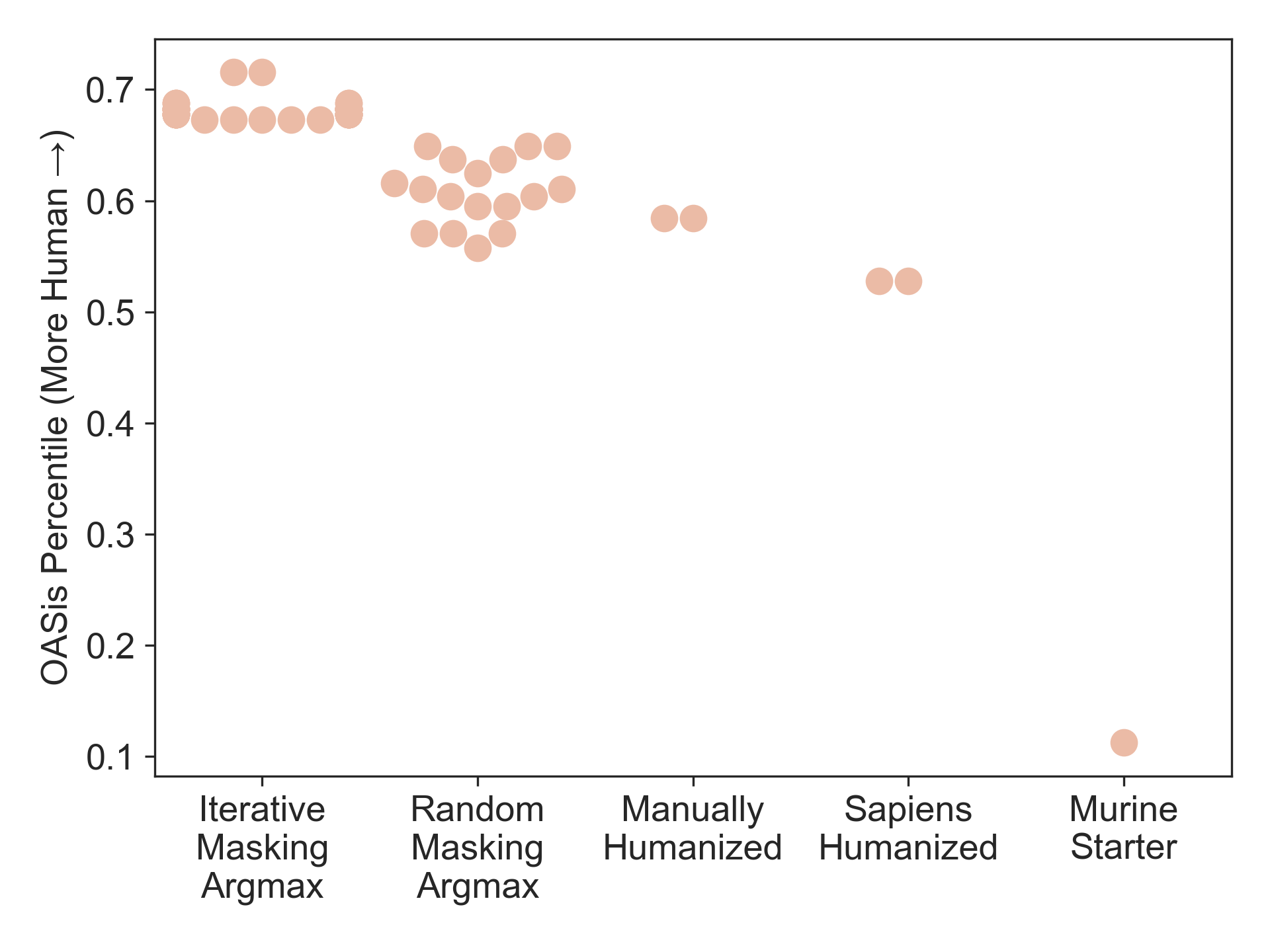}
        \caption{\scriptsize Humanness assessment}
        \label{fig:fig2_labval_humanness}
    \end{subfigure}
    \begin{subfigure}[b]{0.22\textwidth}
        \centering
        \includegraphics[width=\textwidth]{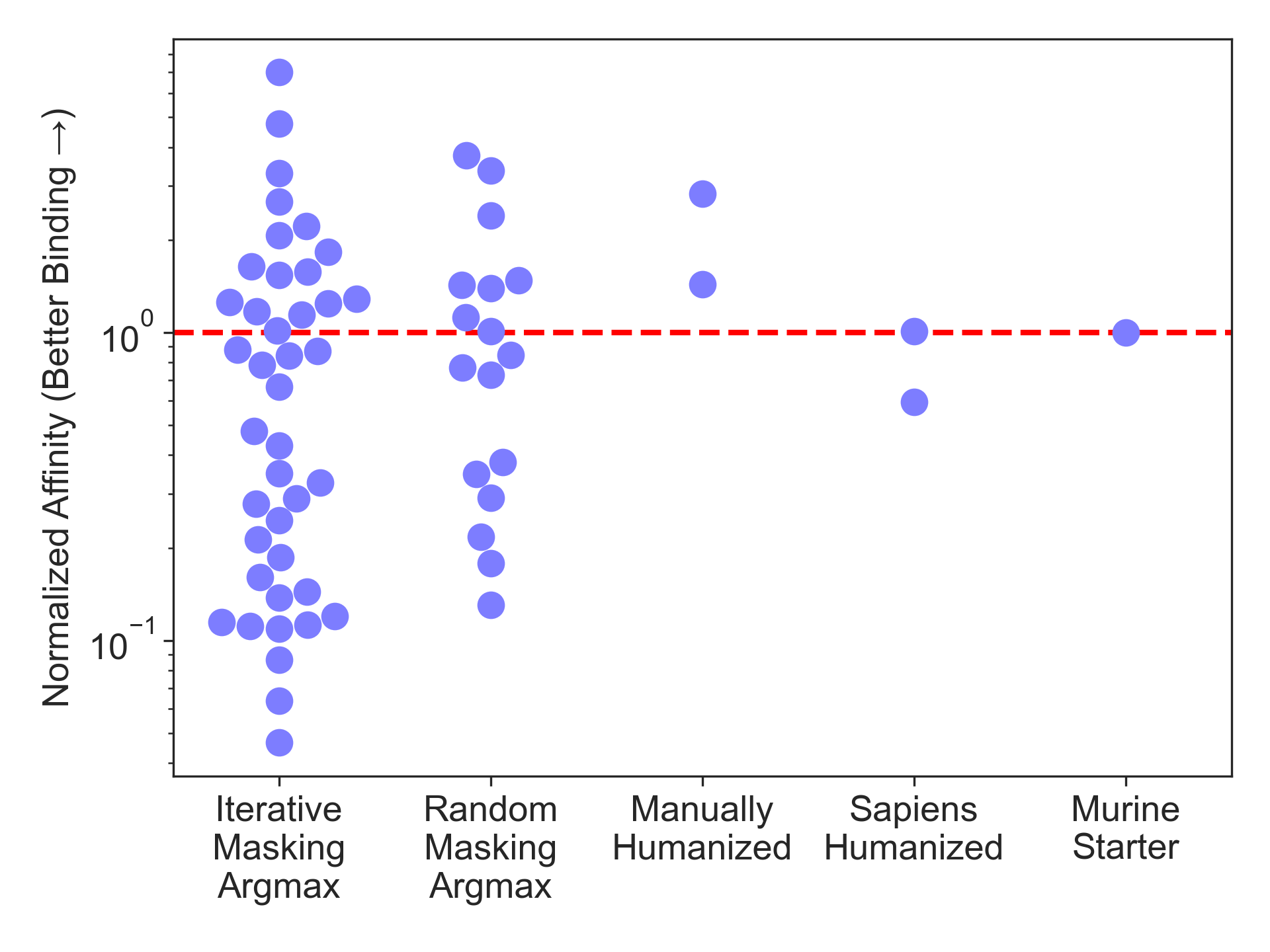}
        \caption{\scriptsize Lab validating affinity}
        \label{fig:fig2_labval_binding}
    \end{subfigure}
    \caption{\small \textbf{Sampling-based approaches generate many candidates with increased humanness} from both murine (a) and manually-humanized (b) starter antibodies. Allowing up to 2 CDR mutations but measuring resulting affinity in the lab gives improved humanness (c) and a range of affinities (d) including some equal to or better than the starter.}
    \label{fig:fig2}
\end{figure}

\cutsubsectionup
\subsection{Sampling Yields Highly Diverse Human Candidates} 
\cutsubsectionup
\label{sec:expts:1}

Our first goal is to understand what impact our sampling-based approaches to humanization have on the number of unique, highly human candidates that can be generated for a given a starting antibody. 

First, we take a set of 10 murine (mouse) clinical antibodies (see Appendix \ref{sec:app_expt_details}) with low starting humanness (average OASis percentile score of 0.06), and humanize them with different strategies, including a manual (germline grafted) baseline -- results are in Figure \ref{fig:fig2_murine}. In all cases humanizing mutations were limited to the framework regions, with the number of mutations ranging from 22 to 35. Sapiens generates only a few, highly-human candidates; in contrast, the masked \texttt{argmax} methods and sampling algorithms all generate orders of magnitude more.

Next, we evaluate a more challenging situation -- further humanization of a set of 30 already-humanized clinical antibodies (average OASis percentile score of 0.41) -- results are in Figure \ref{fig:fig2_human}. On this more difficult task, the sampling-based approaches generate far more highly human candidates as compared to the \texttt{argmax} methods, demonstrating the value of sampling-based humanization for challenging starter molecules.

\textbf{Lab validating binding affinity.} We lab validate a subset of humanized candidates from a murine starter sequence (details in Appendix \ref{sec:app_expt_details}). We synthesize sequences from Random Masking Argmax, Iterative Masking Argmax, Sapiens, HuMAb \citep{marks2021humab} (failed synthesis), and manual humanization, and measure affinities with Bio Layer Interferometry (BLI). Because we have many humanized candidates and will measure affinity in high-throughput, we allow up to 2 humanizing CDR mutations. This gives even more human sequences with a range of affinities including many with equal or better binding (Figure \ref{fig:fig2_labval_humanness}).

\cutsubsectionup
\subsection{Guided Sampling Allows for Joint Iterative Optimization}
We now investigate whether sampling-based humanization can generate candidates with increased humanness that are also enriched for favourable therapeutic properties, such as target binding affinity and thermostability, as part of an iterative antibody optimization campaign. We use the Unmasked Sampling algorithm and use log likelihood as a continuous humanness proxy. For efficiency, we employ a cached oracle approximation, where oracles are first evaluated for all point mutations of the starter sequence and this is used to compute the PoE distribution at each step of sampling. Further details for all experiments are in Appendix \ref{sec:app_expt_details}.

\begin{figure}[t]
    \centering
    \includegraphics[width=\linewidth]{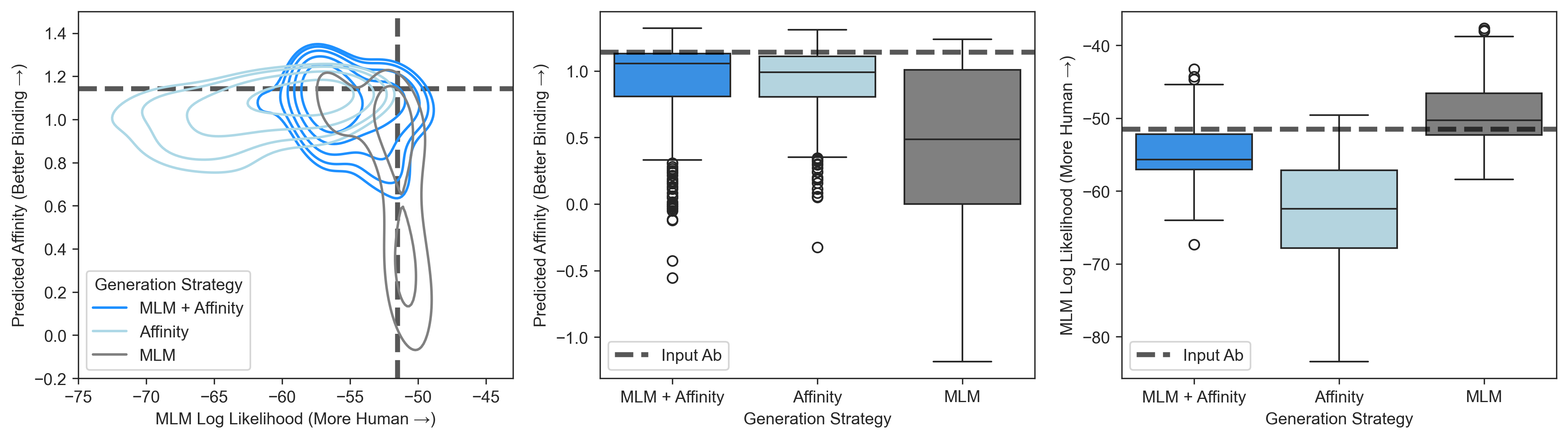}
    \caption{\small \textbf{Guided sampling generates sequences which balance MLM log likelihood ($\sim$ humanness) and predicted target binding affinity} yielding higher-affinity samples compared to unguided sampling (left, contours indicate areas of high sample density, middle and right affinity and log likelihood marginals respectively).}
    \label{fig:aff-guidance}
\end{figure}

\textbf{Single-oracle guidance.} We first study guided sampling with a binding affinity oracle, with results in Figure \ref{fig:aff-guidance}. Guided sampling results in sequences that are enriched for being highly human (high log likelihood) and have high predicted binding affinity.

\begin{figure}[t]
    \centering
    \begin{subfigure}[b]{0.72\textwidth}
        \centering
        \includegraphics[width=\textwidth]{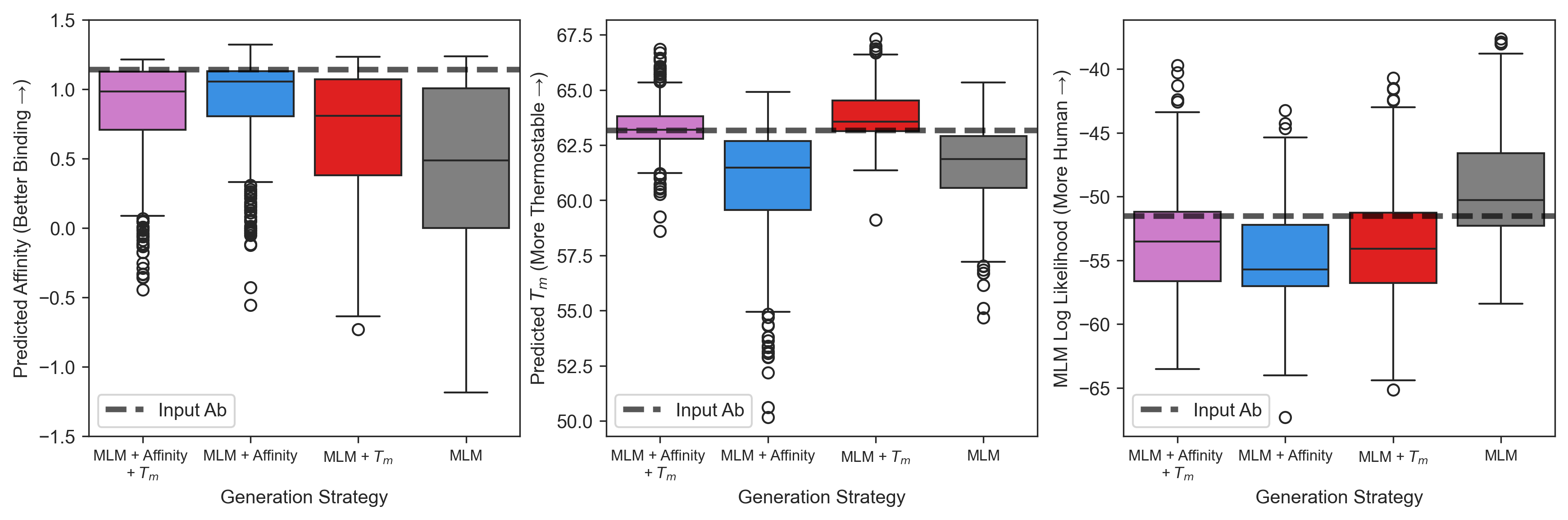}
        \caption{\scriptsize Multioracle Guidance}
        \label{fig:multioracle-guidance}
    \end{subfigure}
    \begin{subfigure}[b]{0.25\textwidth}
        \centering
        \includegraphics[width=\textwidth]{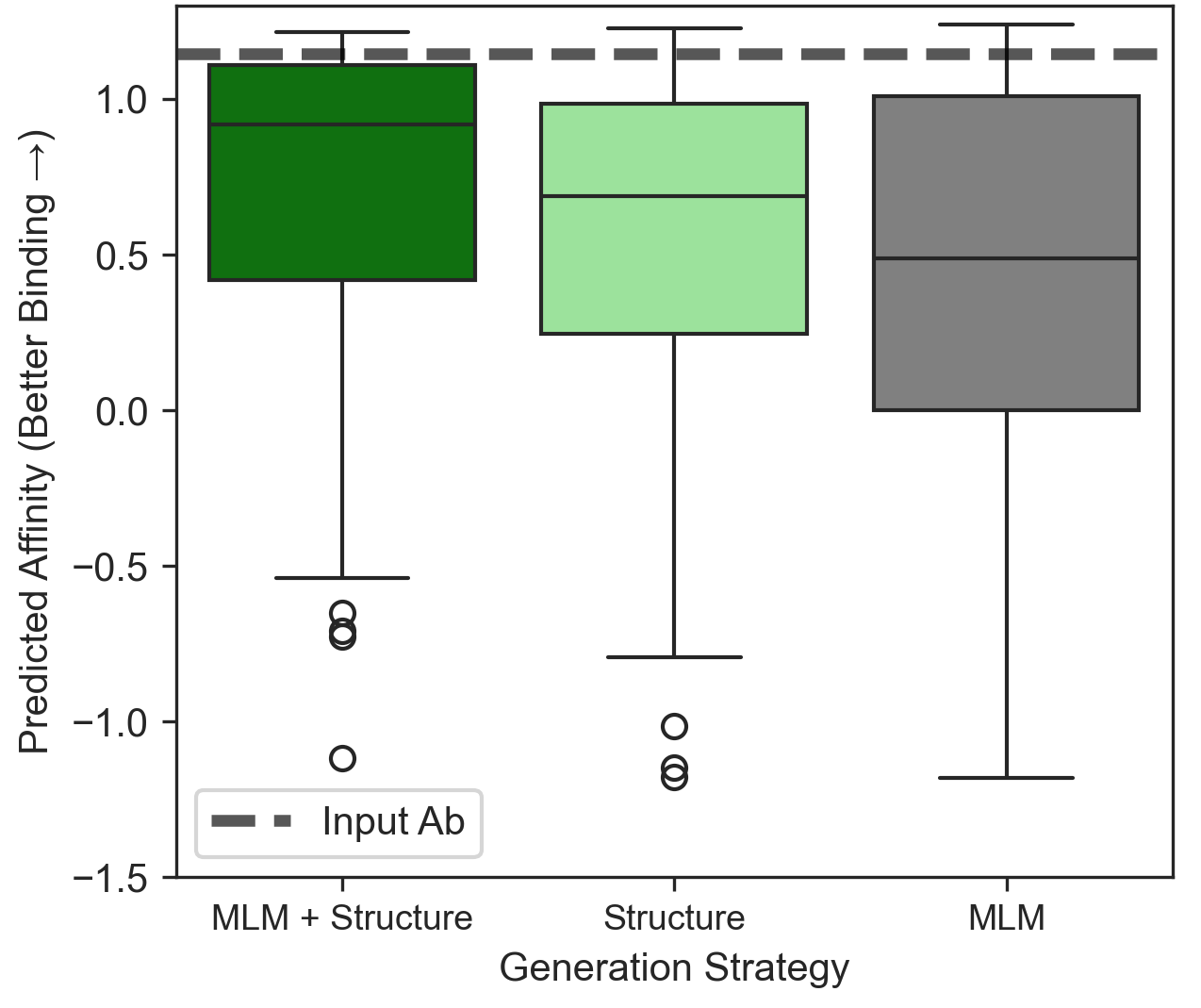}
        \caption{\scriptsize Predicted Affinity with Structure Guidance}
        \label{fig:aff_struct_guidance}
    \end{subfigure}
    \caption{\small \textbf{Guided sampling can flexibly capture multiple desired properties.} Multioracle guidance (a) improves predicted affinity (left), thermostability (middle) and maintains log likelihood (right). Even in the absence of a trained oracle, sampling guided to minimize structural perturbation (b) yields samples which less frequently ablate target binding.}
    \label{fig:complex_guidance}
\end{figure}

\textbf{Multi-oracle guidance.} Extending the above, we generate samples guided by \textit{both} the affinity and thermostability oracles -- Figure \ref{fig:multioracle-guidance} shows the distribution of samples. Multi-oracle sampling results in a greater density of sequences with high log likelihood, predicted affinity, and predicted thermostability, as compared to unguided sampling or when just using one oracle.

In this scenario we are humanizing while iteratively optimizing \textit{e.g.} affinity over several rounds, so we adjust the number and location of mutations introduced accordingly: First, we introduce only a limited number of mutations (up to 6) at each step, and we allow only a subset of these to be within the CDRs (up to 2).

\textbf{Structure-guided sampling.} In situations where we do not have data to train attribute oracles or such oracles are otherwise unavailable, we can use an off-the-shelf antibody structure predictor (e.g., IgFold \cite{ruffolo2022igfold}) in guided sampling to generate candidates that have high log likelihood \textit{and} minimize structural deviation from the starting antibody. These samples are enriched for high predicted binding affinity as compared to unguided sampling (Figure \ref{fig:aff_struct_guidance}), even though an affinity oracle was not used during sampling.

\subsection{Guided Sampling is Effective in Lab Validation} 
We now validate our sampling approaches in the lab, confirming they enrich for sequences with desirable therapeutic properties.

\textbf{Setup.} We evaluate both Guided and Unguided Unmasked Sampling. 
For Guided Unmasked Sampling, we generate 500 candidate sequences that have increased humanness and oracle-predicted binding affinity as compared to the starter. For Unguided Unmasked Sampling, we generate 500 candidate sequences that have increased humanness as compared to the starter. We also included 2 sequences humanized by the Sapiens algorithm. 
Sequences are then filtered for developability liabilities (Appendix \ref{app:guided_sampling_details}).

\begin{wrapfigure}[32]{r}{.45\textwidth}
    \centering    \includegraphics[width=0.42\textwidth]{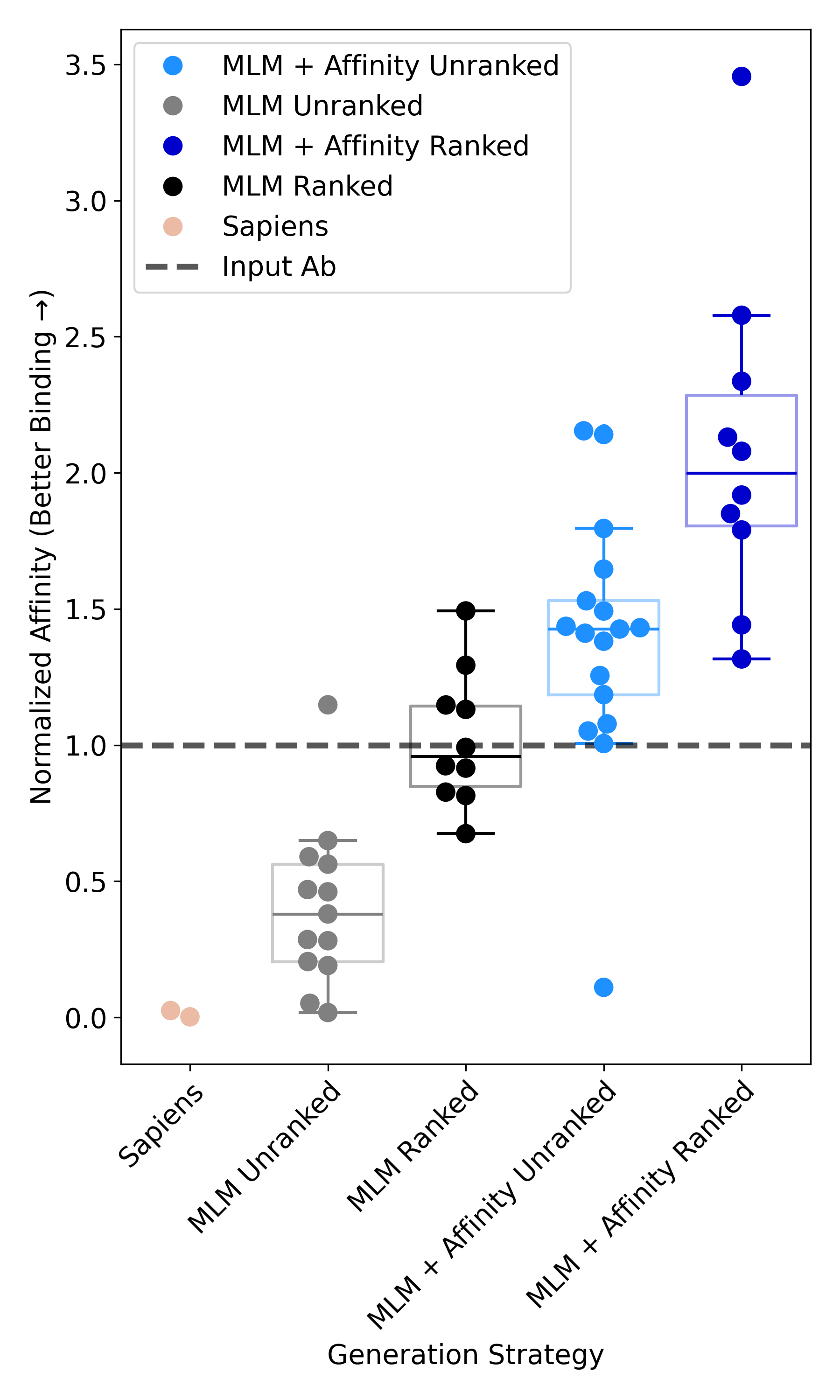}
    \caption{\small \textbf{In lab validation, guided generative humanization yields improved binding affinity.} Affinity guidance outperforms both unguided sampling, as well as unguided samples which are ranked \textit{post-hoc} by an affinity oracle.}
    \label{fig:guided-sample-labvalidation}
\end{wrapfigure}

We study two variants of each method: 
\begin{enumerate}[nosep]
    \item Unranked, where we sample at random 20 sequences.
    \item Ranked, where we rank the sequences with the affinity oracle and then select the top 10.
\end{enumerate} 

The final sets of sequences are then synthesized in the lab and characterized for binding affinity.

\textbf{Results.} Figure \ref{fig:guided-sample-labvalidation} presents the lab validation results -- we see that guided sampling methods outperforms unguided sampling, and ranked guided sampling performs best. The Sapiens humanized sequences failed to bind. This demonstrates the value \textit{in vitro} of biasing the sampling process for predicted high affinity variants, rather than only filtering with an oracle after generation. 

\section{Conclusion}
In this paper, we reframed the problem of antibody humanization as one of conditional sampling from a generative model. This represents a departure from prior work that considered humanization as building a functional mapping between an input antibody and a (often single) humanized output antibody. 
We demonstrate through \textit{in silico} and lab validation that our new sampling based approach can generate many highly human candidate antibodies that are also enriched for properties of therapeutic interest, such as strong binding affinity to a target antigen.

Our method for humanization can be readily embedded into an iterative optimization loop, which could lead to accelerating the development of antibody therapies with reduced immunogenicity risk.

\section*{Acknowledgements}
We would like to thank the BigHat Production team, especially Elisabeth Visser, Caleb Ling, AJ Toledo, and Zach Ormsby and the Data Science team, especially Lindsey Osimiri, Joe Davis, and Dan Anderson. We'd also like to thank Ryan Henrici, Jamie Haddon, and the rest of the BigHat ML team for productive discussions and insightful suggestions.
\clearpage

\bibliography{iclr2024_conference}
\bibliographystyle{iclr2024_conference}
\clearpage
\appendix
\section{Extended Related Work}
\label{sec:app_related_work}
Here, we describe additional related work.

\paragraph{Protein Language Models.} Language models such as BERT or GPT \citep{devlin2018bert, radford2019gpt2} operate over discrete tokenized spaces making them appropriate modeling tools for proteins (also sequences of discrete tokens, namely amino acid residues). There have been many works developing language models specifically for proteins. ESM trained a series of large-scale transformers that were able to learn biological properties of general proteins in an unsupervised manner \citep{rives2021esm, lin2023esm2}. In line with this work, others have followed for antibody-specific protein language models such as AntiBERTy, AntiBERTa, and IgLM \citep{ruffolo2021antiberty,leem2022dantiberta, shuai2023iglm}. Protein language models have been shown to model the evolutionary trajectory of proteins and mature antibodies without conditioning on the antigen \citep{hie2022evolutionary, hie2023efficient}. In this work, we train a protein language model on human antibody data and use it to suggest humanizing mutations to a starter antibody sequence, motivated by the fact that such language models effectively model the human antibody repertoire.

\paragraph{Classical Humanization.} Early work to reduce the immune risk of an antibody involved combining a non-human antibody's variable region with a fully human antibody's constant region, in a process known as grafting. 
In an effort to increase the human content of the antibody even more, many classical humanization methods preserve only the CDRs of the non-human molecule, since these regions are important in antigen binding. In such an approach, a scientist will take antibody sequences found in the human genome, known as germline antibodies, and graft the non-human CDRs onto the human framework. These methods are manual, require a lot of time, and often put other features of the antibody at risk \citep{chirino2004minimizing, harding2010immunogenicity}. Further, often only a single humanized candidate is produced for a single input antibody sequence.

\paragraph{Machine Learning-based Humanization.} As antibody datasets grew in size and methods improved, humanization has become an important machine learning task.
\citet{marks2021humab} train random forest models to classify antibodies as human or nonhuman based on their V gene and uses those models to greedily humanize an antibody using framework mutations. Sapiens trained a masked language model (MLM) on human antibody data, and then used this to suggest humanized sequences by taking the most probable amino acid at each location as determined by the model \citep{prihoda2022biophi}. PLAN uses a protein language model to embed antibody $k$-mers, takes the closest $k$-Nearest-Neighbors, and finally votes on mutations of the initial $k$-mer using the retrieved $k$-mers \citep{zou2023antibody}. Most recently, AbNatiV trains VQ-VAEs on human and camelid antibody data \citep{ramon2023abnativ, oord2017neural}. It then chooses liable positions and suggests mutations at those locations using observed frequency seen at the residues in existing human or camelid samples, while retaining only those that satisfy humanness or camelid-ness criterion. Compared to these methods, our sampling-based humanization approach: (1) generates many more humanized candidates for a starter antibody sequence; and (2) generates highly human candidates that are also enriched for desirable therapeutic properties.

\paragraph{Sampling from Language Models.} Protein language models can provide transition probabilities from one sequence to the next, framing the problem as a Markov random field. If the mutational space has stationary distribution, the problem opens up to Markov Chain Monte Carlo sampling algorithms like Gibbs, Metropolis-Hastings, or the Metropolis-adjusted Langevin algorithm \cite{geman1984stochastic, metropolis1953equation, hastings1970monte, grenander1994mala}. Notably, \cite{wang2019bert} treated the English MLM Bert as a Markov Random Field to generate text using Gibbs sampling. 

When sampling mutations, designers often want multiple different factors to influence sampling. The aforementioned methods only consider a single unsupervised oracle. To overcome this in English language modeling, Plug and Play Language Models (PPLMs) modified the sampling distribution of a model by performing gradient ascent on the final representation of a token using a supervised classifier \cite{dathathri2019plug}. NOS from \cite{gruver2023protein} extended PPLM to diffusion models using both the gradient of supervised classifier and a KL-divergence penalty between the original sequence and the current proposal.

The PPLM line of solutions necessitate a supervised oracle that shares the latent space of the language model of interest, an often untrue assumption in practice and in the context of our experiments. Plug and Play Directed Evolution proposed a general solution to mirror PPLMs, while not requiring gradients \cite{emami2023plug}. The work proposed a Metropolis Hastings algorithm that uses the gradients of the experts to arrive at proposal distributions. With unsupervised $f_i$s and supervised $g_i$s as experts, they viewed log odds of the distribution as $\pi(x) = \log p(x) = \sum_i f_i(x) + \lambda \sum_j g_j(x) - \log Z$, $Z$ being a normalizing constant. They then take the gradients w.r.t. $x$ to create a forward proposal distribution. After proposing $n$ mutations to $x^{(0)}$ making it $x^{(n)}$ the probabilities of mutating from $x^{(n)}$ to $x^{(0)}$ are calculated. A Metropolis-Hastings criteria accepts the sample with probability equal to the product of $\exp(\pi(x^{(n)})-\pi(x^{(0)}))$ and the ratio of the product of forward chain probabilities and reverse chain probabilities.
As another multi-objective sampler, \cite{hie2022high} proposed a programming language for proteins by creating sums of energy functions from different oracles then sampling using an algorithm with qualities of both Metropolis-Hastings and Simulated Annealing.

\clearpage

\section{Additional Methods Details}
\label{sec:app_methods}
\paragraph{Algorithm descriptions.} Algorithms \ref{app:alg:unmasked_sampling}, \ref{app:alg:gibbs_sampling}, and \ref{app:alg:ard_sampling} present Unguided Unmasked Sampling, Unguided Gibbs-like Sampling and Unguided Autoregressive Denoising Sampling respectively.

\begin{algorithm}[h]
    \caption{Unguided Humanization via Unmasked Sampling
    \label{app:alg:unmasked_sampling}}
\begin{algorithmic}[1]
    \STATE {\bfseries Inputs:} starting antibody $\mathbf{x}$, mutation indices $\mathbf{I}$, sampling temperature $\tau$, and an MLM $m$.
    \STATE {\bfseries Output:} a final humanized sequence.
    \STATE 
    \STATE Initialize $\mathbf{x}^{(0)}=\mathbf{x}$ and shuffle $\mathbf{I}$.
    \FOR{$j$ in \texttt{range}(\texttt{len}($\mathbf{I}$))}
        \STATE Compute $\mathbf{Z} = m(\mathbf{x}^{(j)})$
        \STATE Extract row $\mathbf{I}[j]$ from $\mathbf{Z}$, obtaining log probabilities $\mathbf{z}_{\mathbf{I}[j]}$.
        \STATE Sample $r' \sim \texttt{softmax}(\mathbf{z}_{\mathbf{I}[j]}/\tau)$
        \STATE Set $x_{\mathbf{I}[j]}$ = $r'$
        \STATE Set $\mathbf{x}^{(j+1)}=\mathbf{x^{(j)}}$
    \ENDFOR
    \STATE \textbf{Return:} $\mathbf{x}^{(\texttt{len}(\mathbf{I}))}$
\end{algorithmic}
\end{algorithm}

\begin{algorithm}[h]
    \caption{Unguided Humanization via Gibbs Sampling
    \label{app:alg:gibbs_sampling}}
\begin{algorithmic}[1]
    \STATE {\bfseries Inputs:} starting antibody $\mathbf{x}$, mutation indices $\mathbf{I}$, sampling temperature $\tau$, and an MLM $m$.
    \STATE {\bfseries Output:} a final humanized sequence.
    \STATE Initialize $\mathbf{x}^{(0)}=\mathbf{x}$ and shuffle $\mathbf{I}$.
    \FOR{$j$ in \texttt{range}(\texttt{len}($\mathbf{I}$))}
        \STATE Set $x_{\mathbf{I}[j]}$ = \texttt{<MASK>}
        \STATE Compute $\mathbf{Z} = m(\mathbf{x}^{(j)})$
        \STATE Extract row $\mathbf{I}[j]$ from $\mathbf{Z}$, obtaining log probabilities $\mathbf{z}_{\mathbf{I}[j]}$.
        \STATE Sample $r' \sim \texttt{softmax}(\mathbf{z}_{\mathbf{I}[j]}/\tau)$
        \STATE Set $x_{\mathbf{I}[j]}$ = $r'$
        \STATE Set $\mathbf{x}^{(j+1)}=\mathbf{x^{(j)}}$
    \ENDFOR
    \STATE \textbf{Return:} $\mathbf{x}^{(\texttt{len}(\mathbf{I}))}$
\end{algorithmic}
\end{algorithm}

\begin{algorithm}[h]
    \caption{Unguided Humanization via Autoregressive Denoising Sampling
    \label{app:alg:ard_sampling}}
\begin{algorithmic}[1]
    \STATE {\bfseries Inputs:} starting antibody $\mathbf{x}$, mutation indices $\mathbf{I}$, sampling temperature $\tau$, and an MLM $m$.
    \STATE {\bfseries Output:} a final humanized sequence.
    \STATE 
    \FOR{$i$ in $\mathbf{I}$}
        \STATE Set $x_i$ = \texttt{<MASK>}
    \ENDFOR 
    \STATE Initialize $\mathbf{x}^{(0)}=\mathbf{x}$ and shuffle $\mathbf{I}$.
    \FOR{$j$ in \texttt{range}(\texttt{len}($\mathbf{I}$))}
        \STATE Compute $\mathbf{Z} = m(\mathbf{x}^{(j)})$
        \STATE Extract row $\mathbf{I}[j]$ from $\mathbf{Z}$, obtaining log probabilities $\mathbf{z}_{\mathbf{I}[j]}$.
        \STATE Sample $r' \sim \texttt{softmax}(\mathbf{z}_{\mathbf{I}[j]}/\tau)$
        \STATE Set $x_{\mathbf{I}[j]}$ = $r'$
        \STATE Set $\mathbf{x}^{(j+1)}=\mathbf{x^{(j)}}$
    \ENDFOR
    \STATE \textbf{Return:} $\mathbf{x}^{(\texttt{len}(\mathbf{I}))}$
\end{algorithmic}
\end{algorithm}

\clearpage

\section{Additional Experimental Details}
\label{sec:app_expt_details}
\subsection{Dataset details}
\paragraph{Open Antibody Space (OAS).} To train masked language models (MLMs), we use approximately 4 million total unpaired heavy and light chain sequences from the Open Antibody Space \citep{kovaltsuk2018observed} (OAS), a large, unlabelled dataset of antibody sequences. This follows several earlier works \citep{ruffolo2021antiberty, prihoda2022biophi}. We extracted human studies from OAS up to 2019, and derived a training set of 2 million heavy and 2 million light chain antibody sequences, randomly sampling studies until the desired training set size was obtained. 

\paragraph{Oracle training datasets.} The affinity and thermostability oracles were pre-trained on Next-Generation Sequencing (NGS) data from phage display selections containing 1M and 1.9M sequences respectively. Round-over-round enrichments as log-transformed reads-per-million (RPM) ratios were used as regression targets. The affinity models were fine-tuned to 6k BLI (Octet) $K_{D}$ measurements. The thermostability models were fine-tuned to 7k nanoDSF (Uncle) measurements of both $T_{m}$ and $T_{agg}$ as a multi-objective regression task. Fine-tuning data was derived from antibodies produced using Cell-Free Protein Synthesis (CFPS).

\subsection{Model and training details}
\paragraph{Masked language model.} We train a masked language model (MLM) on the OAS subset described above. Our MLM is an 8 layer transformer encoder with 8 attention heads, an embedding dimension of 512, and a feedforward dimension of 2048. It has approximately 25 million trainable parameters in total. This model is trained using a masked language modelling objective, following a similar procedure to Sapiens \citep{prihoda2022biophi} and the original BERT model \citep{devlin2018bert}. During training, we use the same masking configuration as BERT with 15\% of the input amino acid tokens being corrupted overall. Of these, 80\% become \texttt{<MASK>}, 10\% are  randomly corrupted to other residues, and 10\% remain the same. We use train this model for 100 epochs using a cross entropy loss with label smoothing of 0.1 and the AdamW optimizer with a learning rate of 3e-4 and weight decay of 0.01. We use a cosine decay learning rate scheduler with linear warmup for 1000 steps.

\paragraph{Oracle models.} We train two oracle model ensembles. The first predicts an input antibody's binding affinity to a target antigen, $K_{D}$. The second predicts an input antibody's melting and aggregation temperature $T_{m}$ and $T_{agg}$, which is a proxy for the molecule's thermostability (to be developable an antibody must have high thermostability). Note that in the experiments shown here only the $T_{m}$ predictions were used.
Each ensemble consists of 10 models that are trained on different subsets of the input data. At inference time, we obtain a single point prediction for an input antibody by taking the minimum prediction over the antibody (representing a lower bound on the property of interest). Each model in the ensemble is a 1D CNN with a CARP/ByteNet architecture \citep{yang2022convolutions}, modified for regression by removing causal masking and adding a linear projection layer on the output. The affinity and thermostability models had 48 and 64 residual ByteNet blocks respectively and a model dimension of 64, giving a total of 4.1M and 5.5M parameters.

\subsection{Humanness metrics}
\paragraph{OASis Percentile.} The central goal of our work is to increase the `humanness' of antibodies, thereby reducing the immunogenicity risk. As current \textit{in vitro} immunogenicity assays are minimall predictive of immunogenicity in human  patients,  prior work has developed various surrogates to predict immunogenicity, ranging from metrics such as the OASis percentile \citep{prihoda2022biophi}, to models of the underlying biology to predict anti-drug antibody (ADA) response \citep{de2007prediction}. 
We adopt the OASis Percentile from \citet{prihoda2022biophi} since it is is well correlated with ADA response on clinical antibody data and has the best AUROC on the human-non human discriminative task amongst open-source humanness metrics. 
This score is computed by calculating the number of overlapping 9-mer amino acid sequences within an antibody as compared to human 9-mers within the OAS database \citep{kovaltsuk2018observed} at different prevalence levels. This number is normalized and then assigned a percentile based on a set of 544 therapeutic antibodies from the IMGT/mAb-DB database \citep{poiron2010imgt}.

\begin{figure}[t]
    \centering
    \includegraphics[width=0.5\linewidth]{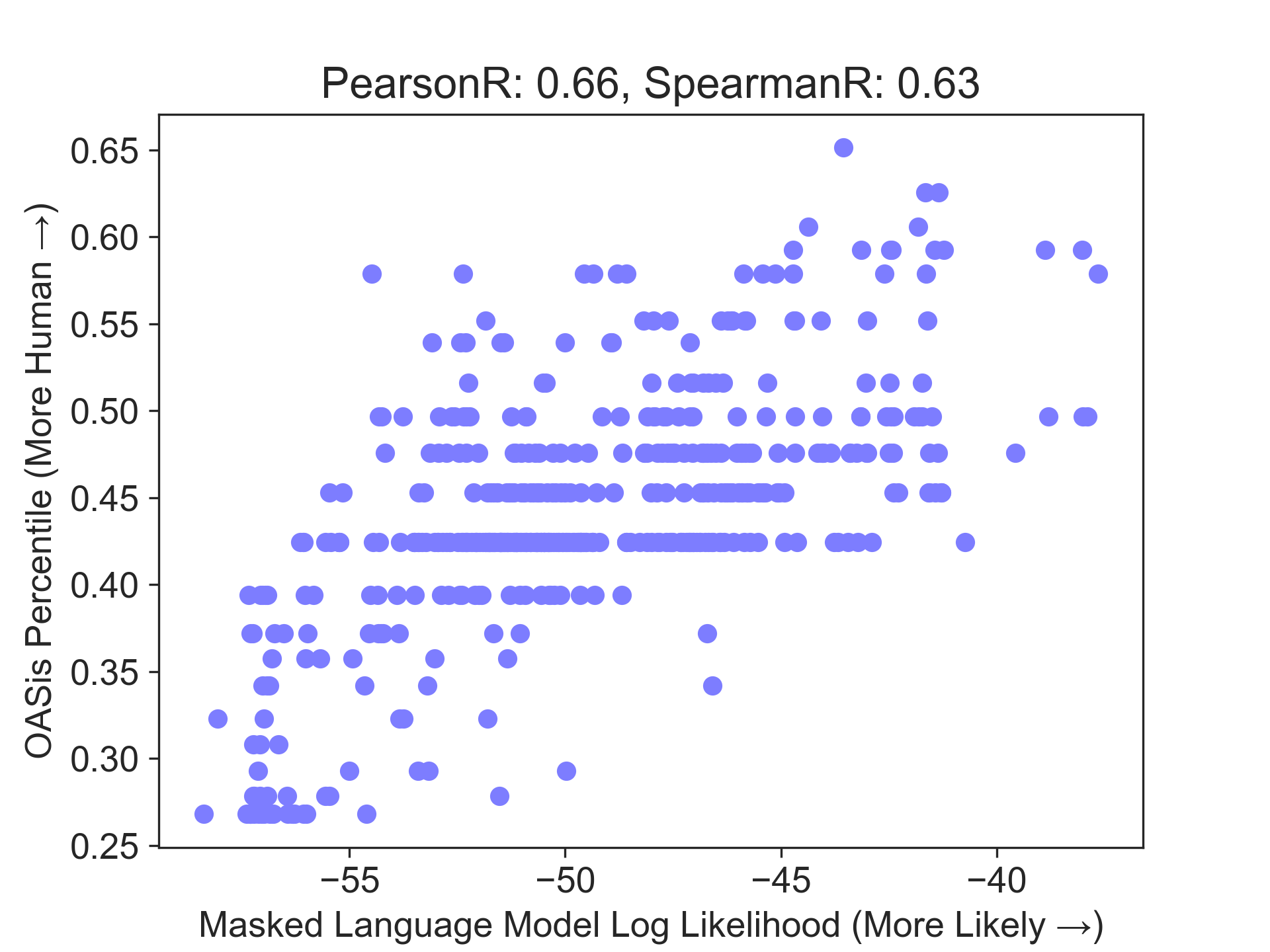}
    \caption{\small Log-likelihood under the MLM is correlated with the OASis percentile score.}
    \label{app:fig:ll_oasis}
\end{figure}

\paragraph{Masked Language Model Log Likelihood.} The OASis percentile score is a discrete quantity and is only indirectly related to the likelihood of a sequence under an OAS language model. Therefore when combining MLMs with other oracles we use the log likelihood of a sequence under our OAS-trained MLM as a measure of that model's contribution to the sampling, as well as a proxy for the humanness of a given antibody. To check that this is a suitable approximation, we compute the OASis score and MLM log likelihood for 500 test sequences and compare them in Figure \ref{app:fig:ll_oasis}. We see a correlation between the two metrics.

\subsection{Additional Details on Unguided Sampling Experiments}
\subsubsection{Experimental Setup} 
\paragraph{Data.} For the \textit{in silico} evaluation the starter murine and humanized antibodies are sampled from a set of existing clinical antibodies taken from \citet{prihoda2022biophi}. We randomly sample 10 murine starters as a simple test case, and 30 humanized starters as a challenging test case. 

For the lab validation, we synthesized the 30 most human random masking argmax framework-region-only humanized versions of a murine starter sequence. Because we are able to validate binding in high-throughput, we also synthesized 230 iterative random masking argmax humanized sequences where we allowed up to 2 mutations in CDRs in addition to framework humanization. HumAb\citep{marks2021humab} and manually humanized forms of the same starter sequence were also included, although the former failed at the synthesis step. Affinities of all synthesizable variants were measured via BLI. This demonstrates that we are able to produce many variants with high humanness, and which are also synthesizable and bind the target (albeit with variable affinity). 

\paragraph{Sample generation hyperparameters.} We set the softmax temperature for the MLM to be equal to 1 for all sampling methods. 
We set all non-CDR locations in the input to be mutable indices (that is, all framework region residues) in order to obtain the most diversity possible, without mutating regions of the sequence that might affect binding.

\subsubsection{Additional results}
In the main text, we showed that all methods except Sapiens generated large numbers of unique samples with increased humanness from the highly non-human murine starter antibodies.
With the partly-humanized starter antibodies, the sampling methods generated many more highly human samples.

\begin{figure}[t]
    \centering
    \begin{subfigure}[b]{0.45\textwidth}
        \centering
        \includegraphics[width=\textwidth]{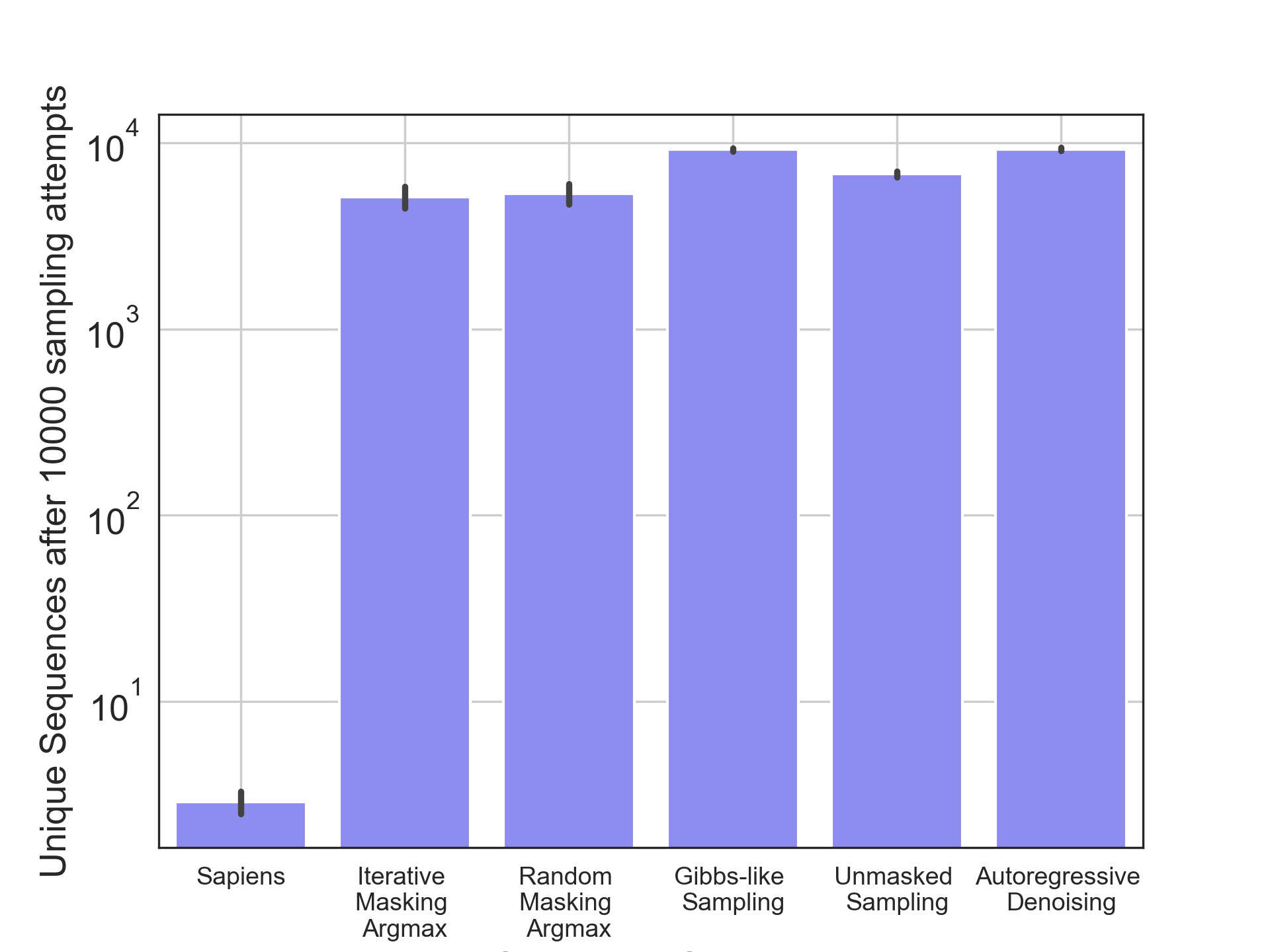}
        \caption{\small Murine Starter}
        \label{app:fig:unique_murine}
    \end{subfigure}
    \hspace{0.05\textwidth}
    \begin{subfigure}[b]{0.45\textwidth}
        \centering
        \includegraphics[width=\textwidth]{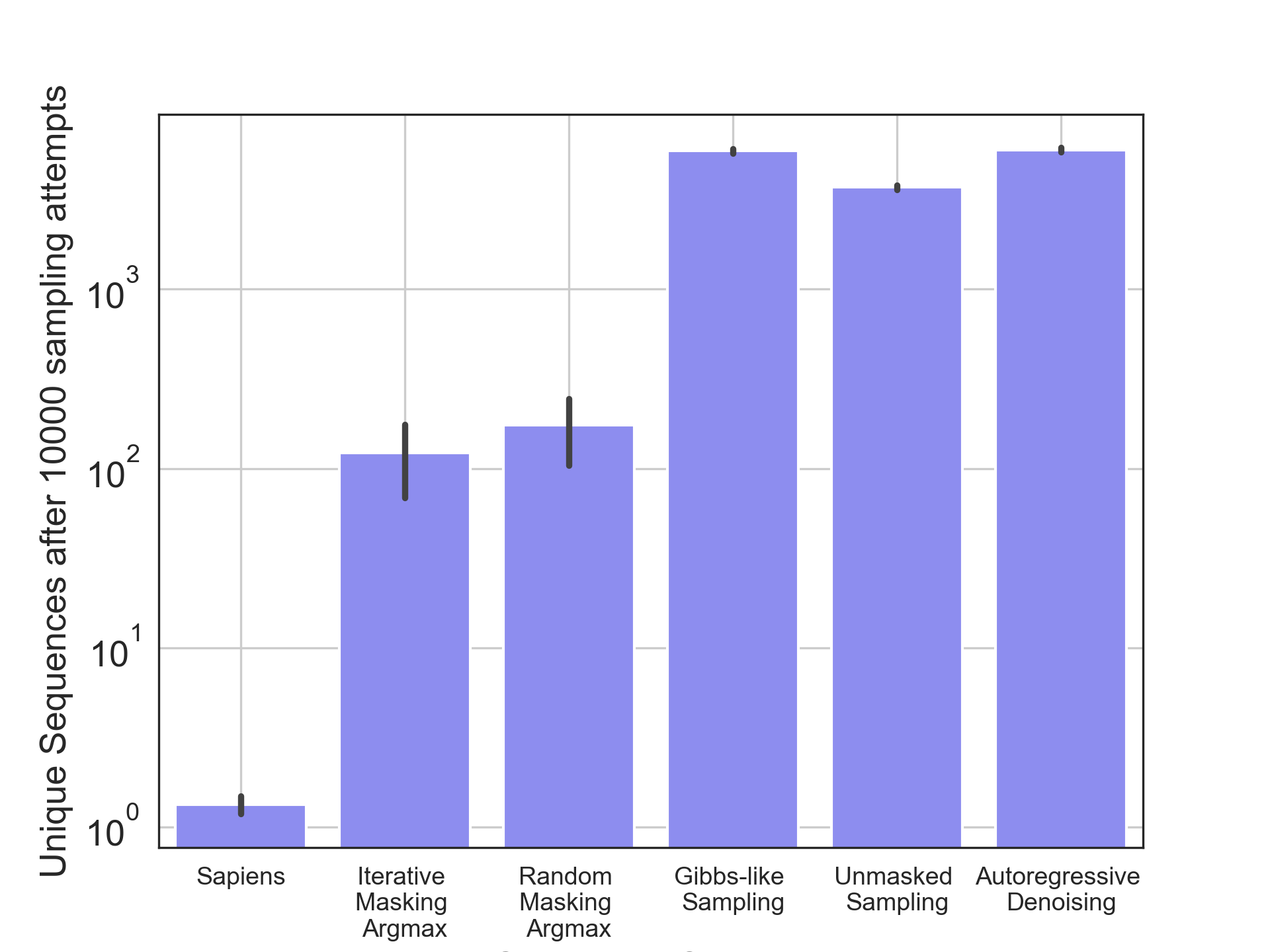}
        \caption{\small Human Starter}
        \label{app:fig:unique_human}
    \end{subfigure}
    \caption{\small \textbf{Sampling-based approaches generate many unique humanized candidates for both Murine and Humanized starter antibodies.}}
    \label{app:fig:num_unique_samples}
\end{figure}

Figure \ref{app:fig:num_unique_samples} additionally presents the raw number of unique samples generated by the different methods for murine starters and humanized starters, without filtering for increased humanness. 
We see that the sampling based techniques produce far more unique sequences overall. 
We observed that for the partly-humanized starters, a large number of generated samples from our the methods have humanness \textit{equal} to that of the starter -- this is perhaps expected, since the starter sequences have already passed through a humanization process, and it is difficult to humanize them further.

\subsection{Additional Details on Guided Sampling Experiments}
\subsubsection{Experimental Setup}
\paragraph{Starter antibody.} We use an already-optimized nanobody (VHH) sequence from an ongoing therapeutic optimization program as the starter sequence. This represents a typical real use-case for humanization, when an antibody sequence with favourable therapeutic properties must undergo a humanization step to reduce the immunogenicity risk.

\paragraph{Sample generation hyperparameters.} We set temperature hyperparameters based on visual inspection of the distribution of predicted affinity and log likelihood of samples. 
When sampling from the MLM with no guidance, we use a temperature of 0.6. 
When sampling from oracle models (affinity, thermostability, structure) without the MLM, we use a temperature of 0.2. 
When sampling with guidance (MLM and one or more oracles) we use an MLM temperature of 1.2 and an oracle temperature of 0.4. 

In these experiments, we choose at random at most 6 locations in the sequence as mutable, with maximally 2 of these being in the CDRs. This allows us to generate a diverse set of samples while also maintaining minimal changes in the CDRs to prevent changes that might affect properties of therapeutic importance such as antigen binding.

For each of the visualizations, we sample 500 unique antibodies from each method. 

\paragraph{Cached oracle approximation.} At every mutable location, our algorithm constructs the product of experts (PoE) sampling distribution by evaluating the oracle for every possible point mutation (20 total) at that location, keeping the rest of the sequence fixed. This evaluation can be computationally intensive, especially with an ensemble of large oracle models. 

As an approximation,  following \citet{emami2023plug}, before running the sampling algorithm,  we evaluate the oracle for all possible single mutations of the starter antibody at every location and store this matrix (for a length 100 antibody, this amounts to 2000 oracle evaluations). 
We then use this pre-computed matrix to obtain the oracle's scores for the different mutations and compute the PoE distribution in \eqref{eq:poe} at each mutable location.
This approximation is exact for mutations with hamming distance 1 (i.e., single point mutations) from the starting sequence. 

\begin{figure}[t]
    \centering
    \includegraphics[width=1\linewidth]{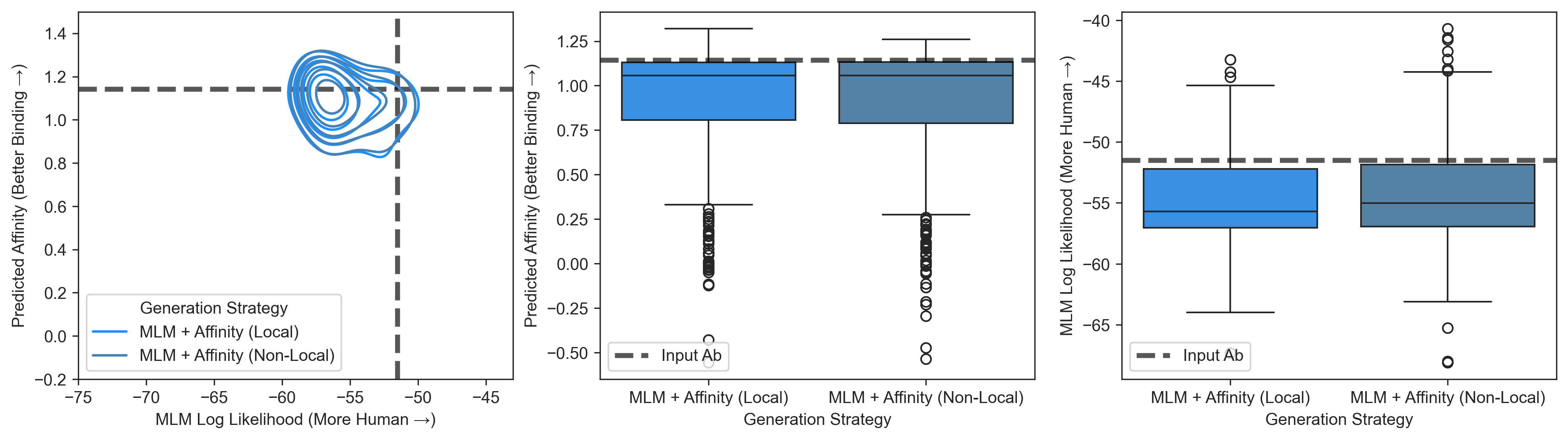}
    \caption{\small \textbf{Using Local PoE results in a similar disribution of generated samples to using Non-Local PoE.}}
    \label{app:fig:local_nonlocal}
\end{figure}

Since we consider only 6 total mutations from the starter, we find that this approximation is effective in practice. Figure \ref{app:fig:local_nonlocal} shows that the distribution of samples with and without this approximation for guided sampling with an affinity oracle, showing that the two distributions are similar.

\subsubsection{Additional results} 
\begin{figure}[t]
    \centering
    \includegraphics[width=1\linewidth]{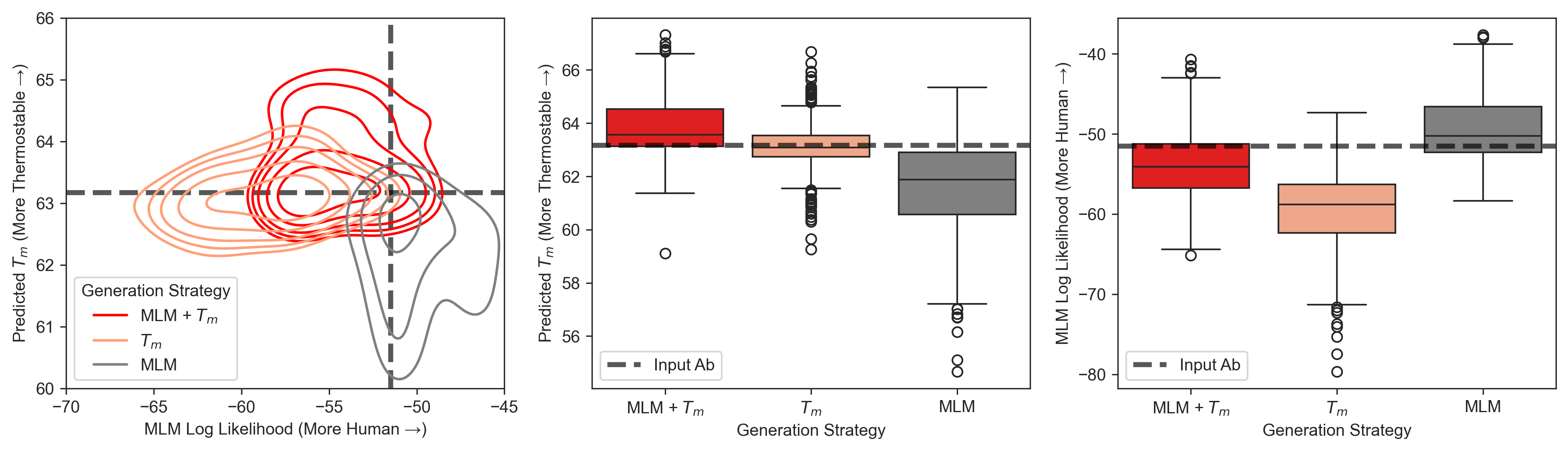}
    \caption{\small \textbf{Guided sampling with a thermostability oracle generates sequences enriched for high melting point.}}
    \label{app:fig:thermo-guidance}
\end{figure}

\paragraph{Guidance with Thermostability Oracle.} In the main paper, we showed how guided sampling with an affinity oracle generates antibodies enriched for high MLM log likelihood (a proxy for humanness) and high oracle-predicted binding affinity. Figure \ref{app:fig:thermo-guidance} presents guided sampling with a thermostability oracle, showing we obtain sequences with high MLM likelihood and high oracle-predicted melting temperature, demonstrating again that our algorithm effectively samples sequences enriched for properties of interest.

\paragraph{Guidance with Structure Prediction Oracle.} As discussed in the main text, there may be situations where attribute prediction oracles are not available for guided sampling. As an example -- early in an antibody optimization campaign, there may not be sufficiently large datasets of antibody sequences and lab-measured attributes to train such an oracle model.
One option in this case is to guide the sampling using an off-the-shelf antibody structure predictor, such as such as IgFold \citep{ruffolo2022igfold} -- this serves as a proxy for similarity to the starter sequence. We now describe this approach.

Given a candidate antibody $\mathbf{x}$, we define: 
$$ S(\mathbf{x}) = \begin{bmatrix}
x_{0_i} & x_{0_j} & x_{0_k} \\
\vdots & \vdots & \vdots \\
x_{{(L-1)}_i} & x_{{(L-1)}_j} & x_{{(L-1)}_k}
\end{bmatrix} \in \mathbb{R}^{L \times 3}$$
to be the antibody's predicted backbone structure. Each row of this matrix represents the 3D coordinates of each alpha carbon atom for each amino acid in the sequence after rigid alignment to the corresponding alpha carbons of the starter antibody  $\mathbf{x}^{(0)}$.

We define the structural score function $f_s$ between the candidate and the starter antibody as follows:
$$ f_s(\mathbf{x}, \mathbf{x}^{(0)}) = - \frac{1}{9L^2}||S(\mathbf{x}) - S(\mathbf{x}^{(0)}) ||_{\text{Fr}},$$
which is the negative of the  Froebenius norm of the difference between the two structures, normalized by dimensionality.
Intuitively, under this definition, a score of 0 represents the highest similarity to the starter antibody, and a highly negative score represents large structural change from the starter.

\begin{figure}
    \centering
    \includegraphics[width=0.9\linewidth]{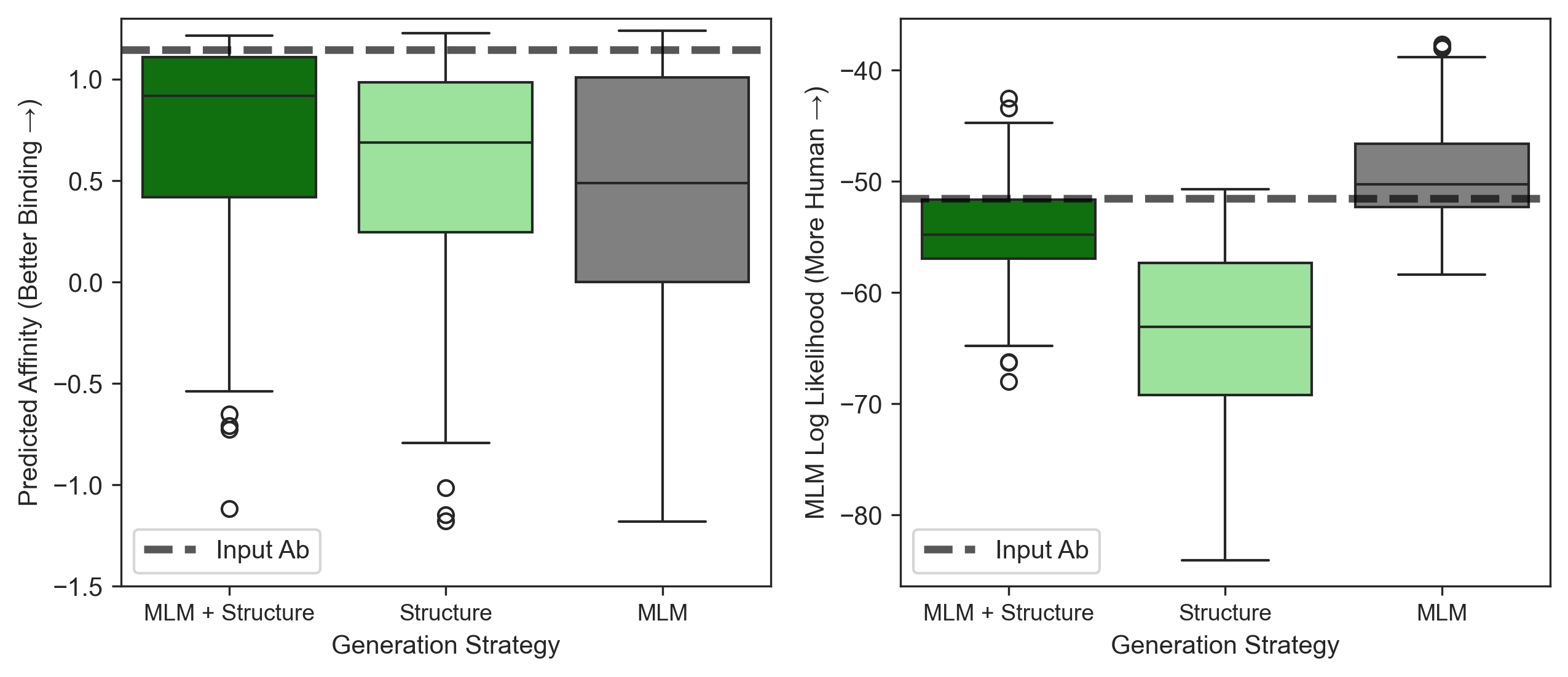}
    \caption{\small Guided sampling with a structural score function results in fewer samples that have poor binding to a target antigen when compared to sampling without guidance (only using the MLM).}
    \label{app:fig:struct_guidance_aff}
\end{figure}

We use this score function to guide sampling, which selects for samples that are structurally close to the starter antibody. We then evaluate the predicted affinity of the sampled sequences using our affinity oracle. 
These results are visualized in Figure \ref{app:fig:struct_guidance_aff}. We observe that incorporating structure guidance results in fewer samples with low predicted binding affinity, suggesting that this score function can help reduce the number of samples obtained that may bind very poorly to the target antigen.

\subsection{Additional Details on Guided Sampling Lab Validation}
\label{app:guided_sampling_details}
\subsubsection{Experimental Setup} 
\paragraph{Data.} We use the same starter antibody as in the previous section. 

\paragraph{Sample generation details.} We use the same sampling hyperparameters and settings as in the previous section.

Humanizing mutations were primarily limited to the framework regions, with only up to 2 humanizing mutations allowed within CDRs (except for Sapiens humanization where CDRs were completely excluded). CDRs were defined as the union of the IMGT, Kabat and Chothia definitions. Sequences were filtered to exclude DDD isomerization and glycosylation motifs, as well as any non-canonical Cysteines.

\paragraph{Wet-lab validation details.} VHHs were synthesized from the humanized sequences via CFPS. Affinities of the variants which were synthesizable in sufficient quantities were measured via BLI (Octet).

%

\end{document}